\documentclass[conference]{IEEEtran}
\IEEEoverridecommandlockouts
\usepackage{cite}
\usepackage{amsmath,amssymb,amsfonts}
\usepackage{algorithmic}
\usepackage{graphicx}
\usepackage{textcomp}
\usepackage{xcolor}
\usepackage{tabularx}
\usepackage{bbm}
\usepackage{booktabs}
\usepackage{multirow}
\usepackage{makecell}
\usepackage{hhline}
\usepackage{threeparttable}
\usepackage{enumitem}
\usepackage[linesnumbered,ruled,vlined]{algorithm2e}
\graphicspath{ {./figs/} }

\newcolumntype{C}[1]{>{\centering\arraybackslash}m{#1}}
\def\BibTeX{{\rm B\kern-.05em{\sc i\kern-.025em b}\kern-.08em
    T\kern-.1667em\lower.7ex\hbox{E}\kern-.125emX}}
\begin{document}

\title{Unsupervised Annotation of Phenotypic Abnormalities via Semantic Latent Representations on Electronic Health Records}

\author{\IEEEauthorblockN{Jingqing Zhang, Xiaoyu Zhang, Kai Sun, Xian Yang, Chengliang Dai, Yike Guo} 
\IEEEauthorblockA{Data Science Institute, Imperial College London, 
London, SW7 2AZ, UK \\
\{jingqing.zhang15, x.zhang18, k.sun, xian.yang08, c.dai, y.guo\}@imperial.ac.uk}
}

\maketitle

\begin{abstract}
The extraction of phenotype information which is naturally contained in electronic health records (EHRs) has been found to be useful in various clinical informatics applications such as disease diagnosis. However, due to imprecise descriptions, lack of gold standards and the demand for efficiency, annotating phenotypic abnormalities on millions of EHR narratives is still challenging. In this work, we propose a novel unsupervised deep learning framework to annotate the phenotypic abnormalities from EHRs via semantic latent representations. The proposed framework takes the advantage of Human Phenotype Ontology (HPO), which is a knowledge base of phenotypic abnormalities, to standardize the annotation results. Experiments have been conducted on 52,722 EHRs from MIMIC-III dataset. Quantitative and qualitative analysis have shown the proposed framework achieves state-of-the-art annotation performance and computational efficiency compared with other methods. 
\end{abstract}

\begin{IEEEkeywords}
Phenotype Annotation, Unsupervised Learning, Natural Language Processing, Deep Learning, Electronic Health Records
\end{IEEEkeywords}

\begin{figure*}[!htb]
\centering
\includegraphics[width=\textwidth]{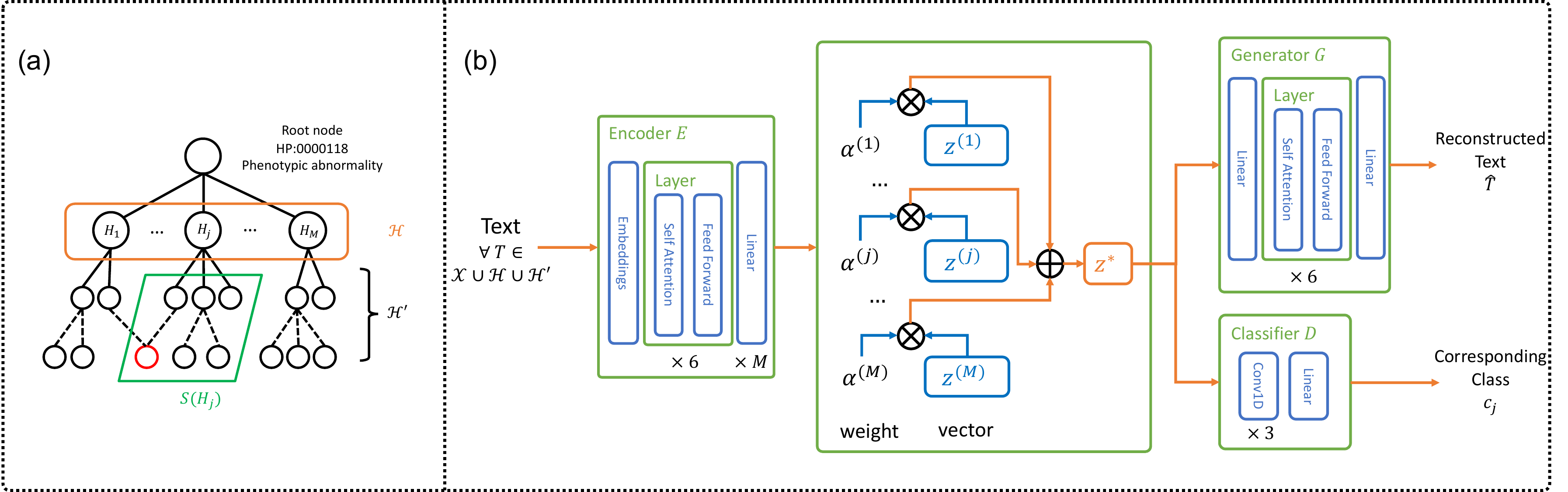}
\caption{\textbf{Left (a):} A simple illustration of Human Phenotype Ontology (HPO). The general phenotypic abnormalities $\mathcal{H}$ in the orange box is what the model aims to annotate from EHRs. Meanwhile, the HPO provides additional subclasses $\mathcal{H}'$. The solid lines show direct relation and the dashed lines show relation with multiple hops. A HPO (red circle) can be a subclass of multiple $H_j\in\mathcal{H}.$ The $S(H_j)\subseteq\mathcal{H}'$ in the green box stands for the additional subclasses of $H_j$. \textbf{Right (b):} The proposed deep learning framework. The textual data is first encoded into the latent space by the encoder $E$. The generator $G$ then reconstructs the textual data so that the latent vectors can adequately represent the semantics. Besides, the classifier $D$ is used to constrain and differentiate the priors. }
\label{fig:framework}
\end{figure*}

\section{Introduction} \label{sec:intro}
Electronic health records (EHRs) are the digital version of patients' paper charts, which are real-time and patient-centered. 
With the increasing adoption of EHRs in hospitals \cite{henry2016adoption}, the explosive information archived in EHRs has been exploited and found to be useful in clinical informatics applications \cite{xiao2018opportunities}, such as disease classification \cite{shi2017towards}
and medical image segmentation \cite{mo2018deep}. 

In this paper, we focus on annotating phenotype information, from EHR textual datasets for better disease understanding. 
In the medical text, the word ``phenotype'' refers to deviations from normal morphology, physiology, or behavior \cite{robinson2012deep}. The EHRs serve as a rich source of phenotype information as they naturally describe phenotypic abnormalities of patients in narratives. The annotation of phenotypic abnormalities from EHRs can improve the understanding of disease diagnosis, disease pathogenesis and genomic diagnostics \cite{deisseroth2018clinphen,son2018deep}, which is a large step towards precision medicine \cite{shickel2018deep}. 


Several standardized knowledge bases have been proposed to help clinicians understand phenotype information in EHRs systematically and consistently \cite{hoehndorf2015role}. Human Phenotype Ontology (HPO) \cite{kohler2016human}, which is a standardized and the most widely used knowledge base of phenotypic abnormalities, provides over 13,000 terms. 
As annotating such amount of phenotypic abnormalities from millions of EHRs manually is extremely expensive and impractical, automatic annotation techniques based on natural language processing (NLP) are demanded.

We first analyzed the appearance of phenotype information in EHRs. With the keyword search approach (i.e. exactly matching the name and synonyms of HPO terms) on the EHRs from MIMIC-III \cite{johnson2016mimic}, we found that on average each EHR contained 40.42 HPO terms against 11.74 ICD\footnote{ICD: https://www.who.int/classifications/icd/en/} codes, and 
the number of HPO terms related to a single disease also varied significantly. For example, regarding the disease \textit{subarachnoid hemorrhage}, the number of HPO terms found in related EHRs ranged from 4 to 40. 
As shown in Table \ref{tab:analysis_hpo_ehr}, the phenotype expressions in the EHRs of patient A and B were clearly different, though they were both diagnosed as \textit{subarachnoid hemorrhage}.
These suggested 
the patients who were diagnosed as the same disease could be further classified into different sub-groups for personalized treatment.
However, the keyword search method cannot maximally extract HPO terms from free text, so more sophisticated automatic phenotype annotation methods is needed.

\begin{table}[htb!]
    \centering
    \caption{Analysis of phenotype information in EHRs from MIMIC-III.}
    \begin{tabular}{|C{0.35\columnwidth}|C{0.55\columnwidth}|}
    \hline
         \textbf{Disease} & \textbf{Phenotype} \\ \hline
         11.74 ICD per EHR &  40.42 HPO per EHR \\ \hhline{|=|=|} 
         \textbf{Disease name} & \textbf{Phenotype quoted from EHRs} \\ \hline
         \multirow{2}{*}{Subarachnoid hemorrhage} & Patient A: ``mild confusion", ``aneurysm" and ``vertebral basilar junction" \\ \cline{2-2}
         & Patient B: ``neurologically stab", ``mild headache" and ``pain is well controlled" \\ \hline
    \end{tabular}
    \label{tab:analysis_hpo_ehr}
\end{table}

There are many automatic annotation methods being developed.
Information retrieval based approaches such as OBO Annotator \cite{taboada2014automated}, NCBO Annotator \cite{jonquet2009ncbo}, Bio-LarK \cite{groza2015human} and MetaMap \cite{aronson2010overview} rely on indexing and retrieval techniques which require manually defined rules and can be computationally inefficient, while deep learning based models with supervision are effective but a gold standard for training is hard to acquire \cite{gehrmann2018comparing} .
However, the problem of how to automatically annotate phenotypic abnormalities from EHRs accurately and efficiently is still far from being solved. First, the phenotypic abnormalities may not be explicitly mentioned in EHRs and the imprecise descriptions in EHR narratives such as abbreviations and synonyms can also make the annotation process difficult. In addition, the reliability of methods is critical in the medical area but it can be difficult to verify on a large-scale dataset due to the cost of collecting phenotype annotations from experts. 

In this work, we propose a novel unsupervised deep learning framework to annotate phenotypic abnormalities from EHRs. Without using any labelled data, the framework is designed to integrate human curated phenotype knowledge in HPO (Figure \ref{fig:framework} (a)).
It is assumed that the semantics of EHRs is a composition of the semantics of phenotypic abnormalities. Based on this assumption, an auto-encoder model and a classifier
are constructed to learn and constrain the semantic latent representations of EHRs. The goal is to learn which phenotypic abnormalities are semantically more important in EHRs. 
The overall structure of the framework is shown in Figure \ref{fig:framework} (b).  The main contributions of this work:

\begin{itemize}
    \item We propose a novel unsupervised deep learning framework to exploit supportive phenotype knowledge in HPO and annotate general phenotypic abnormalities from EHRs semantically. 
    \item We demonstrate that our proposed method achieves state-of-the-art annotation performance and computational efficiency compared with other methods.
\end{itemize}

In the remainder of this paper, we first summarize related works in section \ref{sec:related}. The problem is formalized in section \ref{sec:problem}. We explain the methodology and the deep learning framework in section \ref{sec:method}. The experiments are introduced in section \ref{sec:experiments} and the paper is concluded in section \ref{sec:conclusion}.

\section{Related Works} \label{sec:related}
\subsection{Biomedical Concepts Annotation}
There are many well-established knowledge bases in the medical area such as International Classification of Disease (ICD), Human Phenotype Ontology (HPO) \cite{kohler2016human}, Online Mendelian Inheritance in Man (OMIM) \footnote{OMIM: https://www.omim.org}.
Previous works mostly use indexing and retrieval techniques. For example, the OBO Annotator \cite{taboada2014automated} uses linguistic patterns to retrieve relevant data and then annotates textual snippets based on the indexes of concepts from knowledge bases. Other annotators such as NCBO Annotator \cite{jonquet2009ncbo}, Bio-LarK \cite{groza2015human} and MetaMap \cite{aronson2010overview} follow a similar annotation pipeline. However, those methods suffer from the problem of computational inefficiency. Meanwhile, the evaluation was conducted on limited medical documents and none of the methods was evaluated on EHRs. The recent work \cite{gehrmann2018comparing} shows the effectiveness of supervised deep learning models (CNN) on annotating 10 phenotypes on 1,610 EHRs. In contrast, we propose an unsupervised deep learning framework which is more effective and efficient than the previous works, and our experiments were conducted on 52,722 EHRs.

\subsection{Semantic Representations in NLP}
Learning semantic representations in the latent space of textual data is one of the most fundamental techniques of deep learning in natural language processing. From word embedding \cite{mikolov2013efficient} to sentence encoding \cite{dong2017i2t2i}, the generated latent vectors should adequately represent semantics of text. Without labelled data, learning the representation of semantics in text and supportive knowledge in knowledge bases can be essential to measure semantic similarity and difference \cite{zhang2019integrating}. Our work adopts the idea of using prior distributions to constrain the latent space in generative models \cite{shen2017style} but we aim to annotate the phenotypic abnormalities from EHRs.

\section{Problem Formulation} \label{sec:problem}

There are two types of data sources. First, let $\mathcal{X} = \{X_1, ..., X_N\}$ be a collection of EHRs and each EHR consists of textual notes written by clinicians. 
Second, let $\mathcal{H} = \{H_1, ..., H_M\}$ be a standardized general category of human phenotypic abnormalities provided by HPO. The HPO also provides additional subclasses 
which are notated as $\mathcal{H}'$. A simple illustration of HPO is given in Figure \ref{fig:framework}(a). Each general phenotypic abnormality and subclass comes with a name and a short description. As each $X_i$ is textual data, to comply with this data format, both $\forall H_j \in \mathcal{H}$ and $\forall H'\in\mathcal{H}'$ refer to the textual descriptions of phenotypic abnormalities. 
 
The EHR can include either multiple phenotypic abnormalities or a single or none. Therefore, learning the annotation of phenotypic abnormalities from EHRs is essentially learning the conditional probability $p(\mathbbm{1}_{H_j}|X_i),$ 
$\forall H_j \in \mathcal{H}$, i.e. a binary classification for each $H_j$ to decide whether $H_j$ is mentioned in $X_i$. As a whole, it is a multi-label classification on $\mathcal{H}$.

\section{Methodology} \label{sec:method}

\subsection{Semantic Latent Representations} \label{sec:latent}
To represent the semantics of an EHR $X_i$ and a phenotypic abnormality $H_j$ by latent vector space, the following two assumptions are made:
\begin{enumerate}
    \item The general phenotypic abnormality $H_j$ can be represented by and generated from a latent vector $z^{(j)}_H$ which is sampled from some prior distribution $p(Z^{(j)}_H)$. Each $H_j$ corresponds with a prior distribution $p(Z^{(j)}_H)$ and the prior distributions should be `distinct' enough with each other to highlight their difference. 
    \item The EHR $X_i$ can be represented by and generated from a latent vector $z^*_i$. It is also assumed that the semantics of $X_i$ is a composition of the semantics of  $\forall H_j\in\mathcal{H}$, so $z^*_i=\sum_{j=1}^M\alpha^{(j)}_i z^{(j)}_i$, where $\alpha^{(j)}_i \in [0,1]$ is a weight that can be interpreted the importance of $H_j$ in the composition of $X_i$, and $z^{(j)}_i$ is a sample from the prior distribution $p(Z^{(j)}_H)$ defined above.
\end{enumerate}

Based on the two assumptions made above, it can be noticed that there are two fundamental constraints which should be considered in modelling. 

\begin{enumerate}
    \item The latent vector $z^*_i$ should adequately represent the semantics of $X_i$. Likewise, $z^{(j)}_H$ should adequately represent the semantics of the corresponding $H_j$ (see section \ref{sec:autoencoder}).
    \item The $z^{(j)}_H$ and $z^{(j)}_i$ are both samples from the same prior $p(Z^{(j)}_H)$ and the priors of different $\forall H_j\in\mathcal{H}$ should be `distinct' enough from each other (see section \ref{sec:priors}).
\end{enumerate}

\subsection{An Auto-encoder Model} \label{sec:autoencoder}

Aforementioned, since the annotation process is essentially learning the conditional probability $p(H_j|X_i)$ and the latent space is constructed:

\begin{equation} \label{equ:phx}
\begin{split}
    p(H_j|X_i) &= \int p(H_j, z^*_i | X_i) dz^*_i  \\
               &= \int p(H_j|z^*_i) p(z^*_i|X_i) dz^*_i  \\
               &= \mathbb{E}_{Z\sim p(z^*_i|X_i)} [p(H_j|Z)] \\
\end{split}
\end{equation}

The Equation \ref{equ:phx} suggests an auto-encoder model which is also effective to learn the latent representations \cite{shen2017style}. 
Therefore, a general reconstruction process for all available textual data $T \in \mathcal{X}\cup\mathcal{H}\cup\mathcal{H}'$ is considered.
Both the encoding step and generating step are approximated by deep neural networks. The encoding step $E: \mathcal{T} \rightarrow \mathcal{Z}^*$ and the generating step $G: \mathcal{Z}^* \rightarrow \mathcal{T}$, where $\mathcal{T}$ is the textual space and $\mathcal{Z}^*$ is the latent space. The estimation of parameters of $E$ and $G$ is the optimization of the following.

\begin{equation} \label{equ:autoencoder}
\begin{split}
    &\max \mathbb{E}_{Z\sim p(z^*|T)} \big[ p(T|Z) \big] \\ 
    = &\max_{\theta_E, \theta_G} \mathbb{E}_{Z\sim p_E(T;\theta_E)} \big[ p_G(Z;\theta_G) \big]
\end{split}
\end{equation}

where $\theta_E$ and $\theta_G$ are the parameters of $E$ and $G$ respectively. An illustration of $E$ and $G$ is shown in Figure \ref{fig:framework}(b).


There are three reconstruction loss functions being considered while $E$ and $G$ are estimated. The first loss function considers the general reconstruction loss of EHRs \textit{i.e.} $\forall T=X_i\in \mathcal{X}$.

\begin{equation} \label{equ:x_rec}
    \begin{split}
        \mathcal{L}^{X}_{\text{rec}} = \frac{1}{N}\sum_{i=1}^{N}\bigg[-\log p_G(X_i|E(X_i))\bigg]
    \end{split}
\end{equation} 

Besides, the reconstruction loss of the general phenotypic abnormalities \textit{i.e.} $\forall T=H_j\in\mathcal{H}$ can be defined similarly, and for each phenotypic abnormality $H_j$, the corresponding $\alpha^{(j)}$ should be maximized to 1 and others ($\alpha^{(\neq j)}$) should be minimized to 0, i.e. $z^{(j)}_H=E(H_j)$.

\begin{equation}\label{equ:h_rec}
    \begin{split}
        & \mathcal{L}^{H}_{\text{rec}} = \frac{1}{M}\sum_{j=1}^M\bigg[-\log p_G(H_j|E(H_j)) \\
        &+ \frac{1}{M}\big[-\log(\alpha^{(j)})-\sum_{\substack{k\neq j \\ k=1}}^{M}\log(1-\alpha^{(k)})\big]\bigg]
    \end{split}
\end{equation}

The two loss functions are theoretically sufficient to learn the latent representation of EHRs $\mathcal{X}$ and phenotypic abnormalities $\mathcal{H}$. However, in practice, the short description of $H_j$ may not be informative enough to define all cases of the general phenotypic abnormality, and the usage of the additional subclasses $\mathcal{H}'$ can help the model better understand the general phenotypic abnormalities. Therefore, the reconstruction of the additional subclasses $\mathcal{H}'$ \textit{i.e.} $\forall T=H'\in\mathcal{H}'$ is also necessary and the third loss can be defined as:

\begin{equation}
    \begin{split}
        \mathcal{L}^{H'}_{\text{rec}} =\frac{1}{|\mathcal{H}'|}&\sum_{H'\in\mathcal{H}'}\bigg[-\log p_G(H'|E(H')) \\
        + \frac{1}{M}\big[&-\sum_{\substack{j=1\\ H' \in S(H_j)}}^{M}\log(\alpha^{(j)}) 
        - \sum_{\substack{j=1 \\ H' \notin S(H_j)}}^{M}\log(1-\alpha^{(j)})\big]\bigg]
    \end{split}
\end{equation}
where $S(H_j)\subseteq \mathcal{H}'$ stands for the additional subclasses of the individual $H_j\in\mathcal{H}$. As shown in Figure \ref{fig:framework} (a) (the red circle), there $\exists H'\in\mathcal{H}'$ can be a subclass of multiple $H_j\in\mathcal{H}$. In other words, $\exists k \neq j, H_k,H_j\in\mathcal{H}, S(H_k)\cap S(H_j)\neq \emptyset$.


\begin{algorithm}
\SetAlgoLined
\small
\caption{The training algorithm.}
\label{algo:training}
\KwIn{EHRs $\mathcal{X}$ (training set), general phenotypic abnormalities $\mathcal{H}$, and additional subclasses $\mathcal{H}'$.}
 Initializing $\theta_E$, $\theta_G$, $\theta_{D}$ \;
 \Repeat{convergence} {
    Sample a mini-batch of $B$ textual examples $\{T_{(i)}\}^{B}_{i=1} \subseteq  \mathcal{X}\cup\mathcal{H}\cup\mathcal{H}'$ \;
    Get $z^*_{(i)}$ and $\{z^{(j)}_{(i)}\}_{j=1}^M$ by $E(T_{(i)})$ \;
    Reconstruct $\hat{T}_{(i)}$ by $G(z^*_{(i)})$ \;
    Calculate $\mathcal{L}^{X}_{\text{rec}}$, $\mathcal{L}^{H}_{\text{rec}}$, $\mathcal{L}^{H'}_{\text{rec}}$ respectively \;
    Classify $z^{(j)}_{(i)}$ by $D(z^{(j)}_{(i)})$\;
    Calculate $\mathcal{L}_{\text{pr}}$ by Equation \ref{equ:prior} \;
    Update $\theta_E$, $\theta_G$, $\theta_D$ by gradient descent on:
    \begin{equation}
    \begin{split}
        \mathcal{L}= \lambda_1\mathcal{L}^{X}_{\text{rec}} + \lambda_2\mathcal{L}^{H}_{\text{rec}} 
        + \lambda_3\mathcal{L}^{H'}_{\text{rec}} + \lambda_4\mathcal{L}_{\text{pr}}
    \end{split}
    \end{equation} \\
 }
\KwOut{The encoder $E$.}
\end{algorithm}


\subsection{Constrained and Distinct Priors} \label{sec:priors}
There are two requirements regarding the priors as mentioned in section \ref{sec:latent}. (1) The latent vectors $z^{(j)}_H$ and $z^{(j)}_i$ (both are the outputs of the encoder $E$) should be both sampled from the same prior $p(Z^{(j)}_H)$. (2) The priors of different $H_j\in\mathcal{H}$ should be `distinct' enough from each other because the semantics of different $H_j$ are believed to be different.

To comply with the first requirement, one way is to apply the idea of the variational auto-encoder which uses a KL-divergence to constrain the latent vectors.
Regarding the second requirement, if the latent vectors sampled from different priors can be classified to different classes, then the priors are thought to be `distinct' enough. 

Therefore, considering both the requirements, a classifier $D$ is proposed (Figure \ref{fig:framework}(b)). The classifier $D$ is designed to conduct single-label classification with candidate classes $\{c_1, c_2 ,..., c_M\}$. The intuition is the individual latent vector $z^{(j)}$ from the encoder $E$ of $\forall T \in \mathcal{X}\cup\mathcal{H}\cup\mathcal{H}'$ should be classified as the corresponding class $c_j$ via $D$. Besides, $z^{(j)}$ and $z^{(k)}$ should be classified as two different classes $c_j$ and $c_k$ ($k\neq j$) respectively. Thus, the loss function to constrain and differentiate priors can be defined as follows. 

\begin{equation} \label{equ:prior}
    \begin{split}
        \mathcal{L}_{\text{pr}} = \frac{1}{\#_T}  \sum_{T}\sum_{j=1}^{M}\bigg[-\log p_{D}(c_j|z^{(j)}\in E(T))\bigg]
    \end{split}
\end{equation}




\subsection{Annotation Strategy}

Since the $\alpha^{(j)}_i$ represents the importance of $H_j$ in the composition of $X_i\in\mathcal{X}$ and $\alpha^{(j)}_i \in [0,1]$, in practice, we use $\alpha^{(j)}_i$ to approximate $p(H_j|X_i)$. A threshold $\tau_j$ is applied to each general phenotypic abnormalities $H_j$ to decide if the $H_j$ is mentioned in $X_i$. If $\alpha^{(j)}_i > \tau_j$, then $X_i$ is annotated with $H_j$. Otherwise, $X_i$ is not annotated with $H_j$. The thresholds $\{\tau_j\}_{j=1}^M$ are hyper-parameters and the value of $\tau_j$ for each $H_j$ is decided based on the distribution of $\{\alpha^{(j)}_i\}_{i=1}^{|\mathcal{X}|}$ in the training set (see section \ref{sec:implement}).



\begin{table*}[t]
\centering
\caption{Qualitative analysis to show the effectiveness of our method in discovering implicit phenotypes from EHRs. 
}
\begin{tabular}{|p{0.1\textwidth}|p{0.3\textwidth}|p{0.18\textwidth}|c|c|c|c|c|} \hline
Disease name & Description in EHR & Target HPO & Keyword & NCBO & OBO & MetaMap & Ours \\ \hline
Subarachnoid hemorrhage & On arrival to {[}**Hospital Name**{]} a CT was obtained which showed subarachnoid blood.                          & HP:0001871 (Abnormality of blood and blood-forming tissues) & $\times$    &  $\times$  & $\times$ &  $\times$    & \checkmark \\ \hline
Mitral valve disorder   & He admits to mild DOE, slightly decreased exercise tolerance and occasional palpitations.                        & HP:0003011 (Abnormality of the musculature)                & $\times$     & $\times$   & $\times$  & \checkmark     & \checkmark \\ \hline
Mitral valve disorder   & Patient presents s/p L orbit exenteration ({[}**Masked**{]}) for a history of basal cell carcinoma in her L orbit. & HP:0000478 (Abnormality of the eye)                        &  $\times$   & $\times$  & $\times$  &  $\times$    & \checkmark \\
\hline
\end{tabular}
\label{tab:case_study}
\end{table*}

\begin{table}[htb!]
\centering
\caption{A comparison of different methods. The \#Records refers to the number of textual records used in the original works. The time was measured by the duration of annotating 52,722 EHRs in inference stage with a single thread Intel i7-6850K 3.60GHz and a single NVIDIA Titan X. 
}
\begin{tabular}{|C{0.2\columnwidth}|C{0.2\columnwidth}|C{0.2\columnwidth}|C{0.2\columnwidth}|} 
  \hline
  Method & \makecell{Available (A) \\ Open source (O)} & \#Records & Time to annotate 52,722 EHRs \\ \hline
  OBO & A, Not O & 515 & 1.0 hour \\ \hline
  NCBO & A, Not O& / & 36.7 hours  \\ \hline
  MetaMap & A, O & / & $\sim$ 22 days \\  \hline
  Bio-LarK & Not A, Not O  & 228 &  / \\ \hline
  CNN \cite{gehrmann2018comparing} & Not A, Not O & 1,610 & / \\ \hline
  Ours & A, O & \textbf{52,722} & \textbf{40.2 min} \\
  \hline
\end{tabular}
\label{tab:statistics_baselines}
\end{table}

\begin{table}[htb!]
\centering
\caption{The performance of annotation results compared with the silver standard. All the numbers are averaged across EHRs in the testing set.}
\begin{tabular}{|C{0.2\columnwidth}|C{0.2\columnwidth}|C{0.2\columnwidth}|C{0.2\columnwidth}|} 
  \hline
  Method & Precision & Recall & F1  \\ \hline
  Random    & 0.5541  & 0.5401 & 0.5108 \\ 
  Keyword   & 0.6732  & 0.4982 & 0.5194 \\
  OBO       & 0.6817  & 0.5917 & 0.5775 \\ 
  NCBO      & 0.6782  & 0.5724 & 0.5659 \\ 
  MetaMap   & \textbf{0.7425}  & 0.5231 & 0.5576 \\  \hline
  Ours      & 0.7113 & \textbf{0.6805} & \textbf{0.6383} \\
  \hline
\end{tabular}
\label{tab:accuracy}
\end{table}


\section{Experiments} \label{sec:experiments}
\subsection{Datasets}
We conducted the experiments based on two datasets. (1) We collected 52,722 discharge summaries as the EHRs from \textbf{MIMIC-III} \cite{johnson2016mimic}. Each EHR also came with disease diagnosis marked by International Classification of Diseases (ICD-9) codes. The EHRs were randomly split into a training set (70\%) and a held-out set for testing (30\%). (2) We downloaded phenotype terms from \textbf{Human Phenotype Ontology (HPO)} \footnote{Downloaded in April 2019.} \cite{kohler2016human}. In HPO, each phenotypic abnormality term has a name, synonyms and a definition. Besides, the HPO also provides the class-subclass relations between phenotypic abnormalities, as shown in Figure \ref{fig:framework} (a). There are 24 general phenotypic abnormalities in HPO, \textit{i.e.}, $M=|\mathcal{H}| = 24$, and there are 13,795 additional subclasses, \textit{i.e.}. $|\mathcal{H'}| = 13795$. The vocabulary size $|\mathcal{V}|$ was limited to 30,000 most frequent words in both datasets and all numbers were excluded.


\subsection{Implementation Details \protect\footnote{Source code: https://github.com/JingqingZ/Semantic-HPO.}}  \label{sec:implement}
The encoder $E$ and generator $G$ were implemented based on Transformer \cite{vaswani2017attention}. The encoder $E$ used a word embedding and a position embedding which were followed by 6 stacked Transformer encoders. The hidden size, intermediate size and number of attention heads were set as 768, 3072 and 12 respectively. As $M=|\mathcal{H}|=24$, the $\alpha$s were then calculated by a dense layer with 24 units and a sigmoid activation function. The latent vectors $\{z^{(1)},...,z^{(M)}\}$ were calculated by 24 dense layers, each of which had 1536 units. 
The structure of $G$ was identical to $E$. 
The classifier $D$ was a CNN with three convolution layers. The convolution layers had filter sizes 8, 4, 2 and number of filters 4, 8, 16 respectively. The subsequent dense layer had 24 units with softmax. All the neural networks were implemented by using PyTorch \cite{paszke2017automatic}.

In Algorithm \ref{algo:training}, the loss functions used the cross entropy and the coefficients are set as $\lambda_1=10, \lambda_2=10, \lambda_3=10, \lambda_4=1$ to balance values. The Adam optimizer \cite{kingma2014adam} was used. 
In the annotation afterwards, the value of each threshold $\tau_j$ was set within the range of 70th- and 95th-percentile of $\{\alpha^{(j)}_i\}_{i=1}^{|\mathcal{X}|}$ in the training set. The training and inference (annotation) processes were run on a single NVIDIA Titan X GPU.

Since some original EHRs from MIMIC-III are lengthy, in practice, EHRs were split into fragments each of which had 32 words. After the encoder $E$ was trained, the 
annotation strategy was performed on each fragment and the aggregated annotations of all fragments were used as the final annotations of the EHR. As the ICD codes were reported in the EHRs at different levels, we used 3-digit level ICD codes when evaluating the annotation results for consistency.

\subsection{Evaluation}
We considered the most influential biomedical annotation tools as baselines for performance comparison and the selection was due to their availability and the experimental settings. For a fair comparison, all the annotation results by the baselines were mapped to sets of general phenotypic abnormalities from $\mathcal{H}$.
\begin{itemize}
    \item \textbf{Random choice}: Each EHR was annotated by the general phenotypic abnormalities $\mathcal{H}$ at random.
    \item \textbf{Keyword search}: We searched the name and synonyms of each specific phenotypic abnormality in all EHRs and used the searching results as the annotations. 
    \item \textbf{OBO Annotator} \cite{taboada2014automated} \footnote{OBO: http://www.usc.es/keam/PhenotypeAnnotation/}: Java implementation.
    \item \textbf{NCBO Annotator} \cite{jonquet2009ncbo} \footnote{NCBO: http://data.bioontology.org/documentation}: the annotator web APIs.
    \item \textbf{MetaMap} \cite{aronson2010overview}: 2016v2. 
\end{itemize}

Since it is impractical to collect a gold standard on thousands of EHRs, we created a silver standard, i.e., a mapping from ICD codes to the general phenotypic abnormalities $\mathcal{H}$ from HPO. As there is no manual curated direct mapping between ICD codes and HPO terms, we used Online Mendelian Inheritance in Man (OMIM), which is a catalog of human genes and genetic disorders, as an intermediate hop to link ICD codes and HPO terms. We collected the mapping from ICD codes to OMIM phenotype entries \cite{goh2007human,park2009impact} and the mapping from OMIM entries to HPO terms \footnote{https://hpo.jax.org/app/download/annotation}. Based on these two manual curated mappings, we constructed a mapping from ICD codes to HPO terms, \textit{i.e.} the general phenotypic abnormalities $\mathcal{H}$. The silver standard of the annotations of $\mathcal{H}$ from EHRs was constructed by using this mapping.

The constructed silver standard provides a rich information source on diseases and their associated phenotypic characteristics. With the silver standard, we can partially evaluate the reliability of the annotation results by different methods. 
We used the micro-precision, micro-recall and micro-F1 which are averaged across EHRs for quantitative analysis. 
Some typical cases where EHRs have implicitly described some phenotypic abnormalities were shown for qualitative analysis. 



\subsection{Results and Discussion}

Table \ref{tab:statistics_baselines} compares different methods to show the scalability and efficiency of our method. Our work has conducted experiments on 52,722 EHRs, which are significantly more than previous works. In addition, our method is also more computationally efficient than the baselines. The inference (annotation) stage of our method takes 40.2 minutes to annotate 52,722 EHRs, which is 33\% faster than the OBO Annotator and $>$98\% faster than the NCBO Annotator and MetaMap.



Table \ref{tab:accuracy} compares the accuracy of different annotation methods on the silver standard. The proposed method achieves the precision of 0.7113, recall of 0.6805 and F1 of 0.6383. The F1 is significantly higher than those of all other baselines. 
Considering the association between phenotype and diseases in the silver standard, we believe that our method is more effective and the annotation results of our method can provide a better indication for disease diagnosis than the baselines.

Along with the evaluation using the silver standard, we have conducted qualitative analysis to provide more insights of our annotation work. The EHRs for qualitative analysis are selected from the patients with single disease. We find that within the same disease group, the phenotypic abnormalities vary across different EHRs. Our method can identify the HPO terms that are missed by other methods.
Table \ref{tab:case_study} shows three typical case studies from qualitative analysis, where one EHR is from the disease  \textit{subarachnoid hemorrhage} and two EHRs are from the disease \textit{mitral valve disorder}. 
In the first case, the EHR contains the keyword ``subarachnoid blood" that clearly indicates the the presence of the general phenotype category ``HP:000187 (Abnormality of blood and blood-forming tissues)", while only our annotation method has found this HPO term. The EHR in the second case describes ``slightly  decreased exercise tolerance" which indicates movement impairment, and both our method and MetaMap have successfully found the related general phenotype category ``HP:0003011 (Abnormality of the musculature)". In the third case, although the EHR is originally diagnosed as \textit{mitral valve disorder}, its description shows that this EHR can be wrongly diagnosed as it is more likely to have eye diseases. Our method has annotated the general phenotype category ``HP:0000478 (Abnormality of the eye)", which is consistent with our manual investigation. From the listed cases, we show that our annotation method outperforms others in the aspect of finding phenotypic abnormalities from implicit information.

\section{Conclusion and Future Work} \label{sec:conclusion}
In this work, we propose a novel unsupervised deep learning framework to annotate phenotypic abnormalities from EHRs. The proposed framework is able to learn semantic latent representations of textual data and use different prior distributions to constrain the latent space. The experiments have shown the effectiveness, efficiency and scalability of our method and we believe our method can provide a better indication for disease diagnosis than the baselines. In the future, we plan to extend the proposed framework to annotate all the 13,000 specific phenotypic abnormalities in HPO. Besides, due to the generality of the proposed framework, we believe it can be applied to annotating general concepts on plain text in general domains if a well-established knowledge base is available.


\section*{Acknowledgment}
Jingqing Zhang would like to thank the support from the LexisNexis\raisebox{1ex}{\small{\textregistered}} Risk Solutions HPCC Systems\raisebox{1ex}{\small{\textregistered}} academic program and Pangaea Data.

\bibliographystyle{IEEEtran}
\bibliography{IEEEfull.bib}

\end{document}